\documentclass[lettersize,journal]{IEEEtran}
\usepackage{amsmath,amsfonts}
\usepackage{algorithmic}
\usepackage{algorithm}
\usepackage{array}
\usepackage[caption=false,font=normalsize,labelfont=sf,textfont=sf]{subfig}
\usepackage{textcomp}
\usepackage{stfloats}
\usepackage{url}
\usepackage{verbatim}
\usepackage{graphicx}
\usepackage{cite}
\usepackage{wrapfig}
\usepackage{color}
\usepackage{colortbl}
\usepackage{makecell}

\usepackage[utf8]{inputenc} 
\usepackage[T1]{fontenc}    
\usepackage{hyperref}       
\usepackage{url}            
\usepackage{booktabs}       
\usepackage{amsfonts}       
\usepackage{nicefrac}       
\usepackage{microtype}      
\usepackage{xcolor}         
\usepackage{tabularx}
\usepackage{graphicx}

\usepackage{multirow}

\newcommand{\etal}{\textit{et al.}}

\newcommand{\Skip}[1] {}

\usepackage{verbatim}

\usepackage{amsthm}
\usepackage{amsmath}
\usepackage{subcaption}

\usepackage{authblk}

\Skip{
\makeatletter
\renewcommand\AB@affilsepx{, \protect\Affilfont}
\makeatother}

\makeatletter 
  \newcommand\figcaption{\def\@captype{figure}\caption} 
  \newcommand\tabcaption{\def\@captype{table}\caption} 
\makeatother

\theoremstyle{definition}
\newtheorem{definition}{Definition}[section]

\begin{document}

\title{Unifying Retrieval and Segmentation: A Hypergraph Diffusion Approach with Spatial Verification and Uncertainty Awareness}

\title{Fast and Accurate Pixel Retrieval via Spatial and Uncertainty Aware Hypergraph Diffusion}

\author{Guoyuan An, Yuchi Huo, Sung-Eui Yoon,~\IEEEmembership{Senior Member,~IEEE}
\thanks{The authors are with the SGVR Group, KAIST. }
\thanks{Manuscript received Mar 01, 2024.}}

\markboth{Journal of \LaTeX\ Class Files}%
{Shell \MakeLowercase{\textit{et al.}}: A Sample Article Using IEEEtran.cls for IEEE Journals}


\maketitle

\begin{abstract}
This paper presents a novel method designed to enhance the efficiency and accuracy of both image retrieval and pixel retrieval. Traditional diffusion methods struggle to propagate spatial information effectively in conventional graphs due to their reliance on scalar edge weights. To overcome this limitation, we introduce a hypergraph-based framework, uniquely capable of efficiently propagating spatial information using local features during query time, thereby accurately retrieving and localizing objects within a database.

Additionally, we innovatively utilize the structural information of the image graph through a technique we term ``community selection''. This approach allows for the assessment of the initial search result's uncertainty and facilitates an optimal balance between accuracy and speed. This is particularly crucial in real-world applications where such trade-offs are often necessary.

Our experimental results, conducted on the (P)ROxford and (P)RParis datasets, demonstrate the significant superiority of our method over existing diffusion techniques. We achieve state-of-the-art (SOTA) accuracy in both image-level and pixel-level retrieval, while also maintaining impressive processing speed. This dual achievement underscores the effectiveness of our hypergraph-based framework and community selection technique, marking a notable advancement in the field of content-based image retrieval.
\end{abstract}

\begin{IEEEkeywords}
Landmark retrieval, segmentation, graph diffusion, spatial verification, retrieval uncertainty
\end{IEEEkeywords}

\section{Introduction}
\label{sec:intro}

Image search, a fundamental task in computer vision, has seen significant advancements even before the advent of deep learning~\cite{radenovic2018revisiting,philbin2008object, chum2007total}. Despite these successes, challenges remain, particularly when the object of interest occupies only a small part of the true positive image, which often includes a complex background. This complexity can hinder users from recognizing the correct images easily, even when the search engine ranks them highly~\cite{an2023towards}. To address this problem, recent studies emphasize the importance of not just identifying but also precisely localizing and segmenting the query object within the retrieved images~\cite{an2023towards,shen2012object}. This new task, known as pixel retrieval, aims to enhance the accuracy and user-friendliness of image search results.

There are several existing approaches for pixel retrieval. SPatial verification (SP) examines the spatial arrangement of matched local features to pinpoint the target object in database images~\cite{philbin2007object}. Meanwhile, detection and segmentation methods treat the query image as a learning example to localize and segment the target in database images~\cite{osokin20os2d, liu2016ssd}. Dense-matching methods involve training neural networks to establish pixel correspondences for query and database image pairs~\cite{truong2021warp}. Although these methods show promise in pixel retrieval performance, they share a common limitation: the real-time computation of pixel correspondences for each image pair, leading to significant time consumption~\cite{an2023towards}.

In real-world applications, pixel retrieval demands both speed and accuracy. Our paper proposes a novel approach to address this need: pre-computing spatial matching information of database images offline and swiftly propagating this information for any given online query. We introduce a spatial aware hypergraph diffusion model, which efficiently constructs a hypergraph for a given query and propagates spatial relationships through hyperedges across a sequence of images.

This novel technique excels in achieving rapid and accurate pixel retrieval, and it also notably enhances the performance of image-level retrieval of relation-based reranking methods like query expansion and diffusion.  Query expansion~\cite{chum2007total,gordo2017end,radenovic2018fine,arandjelovic2012three,gordo2020attention} issues a new query by aggregating useful information of nearest neighbors of the prior query. Unfortunately, this approach only explores the neighborhood of similar images, resulting in a low recall~\cite{iscen2017efficient}. On the other hand, diffusion~\cite{iscen2017efficient,zhou2004ranking} explores more images on a neighborhood graph of the dataset. It, however, adopts false positive items in practice, leading to a low precision~\cite{radenovic2018revisiting}. 

A key issue in these methods is the spatial ambiguity in propagation, as illustrated in Fig.~\ref{fig:graph}. Since each database image may contain multiple objects, a sequence of images identified based on simple global feature similarity might not consistently contain the same object. Our hypergraph diffusion technique addresses these low recall and precision issues by effectively propagating to only corresponding local features along spatially aware hyperedges.

Another contribution of this paper is the development of a novel method for measuring the uncertainty of retrieval results. Predicting retrieval uncertainty is an essential but less explored topic in image retrieval; it is crucial for balancing accuracy and speed in image search engines~\cite{an2021hypergraph}. We leverage the graph structure of the database, wherein each image is a node connected to similar nodes. Our hypothesis is that images in the same community likely contain the same object. A high-quality retrieval for a query would typically yield images from a singular community. Conversely, a spread across diverse communities indicates higher uncertainty, prompting the search engine to employ more intensive reranking techniques. We term this approach ``community selection.''

Our experimental findings demonstrate that the spatial aware hypergraph diffusion method achieves outstanding performance in both image-level and pixel-level retrieval, surpassing current state-of-the-art methods. On the challenging ROxford dataset, our hypergraph-based approach records impressive mean Average Precisions (mAPs) of 73.0 and 60.5, with and without R1M distractors, respectively. Moreover, our method exhibits superior pixel retrieval accuracy at a significantly faster speed compared to existing techniques. Additionally, the community selection technique effectively reflects the actual quality of retrieval, confirming its efficiency.

In summary, our contributions are threefold. 
\begin{enumerate}
    \item We introduce a fast and accurate method for pixel retrieval, offloading the slow spatial matching process offline and leveraging a novel hypergraph model for real-time spatial information propagation. 
    \item We demonstrate that this hypergraph diffusion method substantially improves the image-level retrieval performance of existing relation-based reranking methods, achieving state-of-the-art accuracy on ROxford and RParis datasets. 
    \item We propose a unique community selection strategy to predict retrieval quality without user feedback, effectively balancing accuracy and speed in real-life applications.
\end{enumerate}

\section{Related works}
\label{sec:related}

\subsection{Pixel retrieval}
Pixel retrieval, an advanced variant of image retrieval, concentrates on identifying and segmenting pixels associated with a query object within database images~\cite{an2023towards}. It aims to offer more detailed and user-centric results compared to traditional image retrieval~\cite{shen2012object,an2023towards}.

In the realm of pixel retrieval, existing methods can be classified into spatial verification~\cite{chum2007total,philbin2007object, cao2020unifying}, detection and segmentation~\cite{fan2022ssp, min2021hypercorrelation}, and dense matching~\cite{truong2021warp}. Spatial verification evaluates the spatial location consistency of matched local features, such as SIFT~\cite{lowe2004distinctive}, DELF~\cite{Noh_2017_ICCV}, and DELG~\cite{cao2020unifying}. Originally developed to enhance image-level retrieval performance, it also effectively pinpoints the location of objects in images.

Techniques like Open-World Localization (OWL)~\cite{minderer2022simple}, Hypercorrelation Squeeze Networks (HSNet)~\cite{min2021hypercorrelation}, and the Self-Support Prototype model (SSP)~\cite{fan2022ssp} represent the detection and segmentation approach. They treat the query image as a one-shot example to localize or segment the target object in database images.

Dense matching methods, including prominent ones like GLUNet~\cite{truong2020glu} and PDCNet~\cite{truong2021pdc}, focus on establishing dense pixel correspondences. They play a crucial role in identifying corresponding points in database images for pixels in the query image, thereby achieving pixel retrieval objectives.

However, these methods require identifying corresponding pixels between the database and query images during the query time, a process that can be time-consuming and negatively impact user experience in retrieval applications~\cite{radenovic2018revisiting,an2023towards}. To overcome this limitation, this paper introduces an innovative approach that significantly speeds up the pixel retrieval process while maintaining high accuracy, thereby enhancing efficiency and user experience in pixel retrieval applications.

\subsection{Relation-based search: query expansion and diffusion}

Numerous methods leverage the interrelations among database images to improve search results. These techniques are primarily categorized into query expansion (QE)~\cite{philbin2007object} and diffusion~\cite{zhou2004ranking}. 

QE~\cite{chum2007total} aggregates the top-ranked initial candidates into an expanded query, which is then used to search for more images in the database. The critical factor here is aggregating only helpful information and culling out unrelated information. 
Average query expansion (AQE)~\cite{chum2007total} mean-aggregates the top k retrieved images as the expanded query. Average query expansion with decay (AQEwD)~\cite{gordo2017end} gives the top k images the monotonically decaying weights over their ranks. Alpha query expansion ($\alpha$QE)~\cite{radenovic2018fine} uses the power-normalized similarity between the query and the top-ranked images as the aggregation weights. Discriminative query expansion (DQE)~\cite{arandjelovic2012three} also uses the weighted average, where the weight is the dual-form solution of an SVM, which classifies the positive (top-ranked) and negative (low-ranked) images of a query. Learnable attention-based query expansion (LAttQE)~\cite{gordo2020attention} uses self-attention~\cite{vaswani2017attention} to share information between the query and the top-ranked items to compute better aggregation weights.

Diffusion~\cite{zhou2004ranking,iscen2017efficient} ameliorates the abovementioned problem by propagating similarities through a pairwise affinity matrix. The works of Chang~\etal~\cite{chang2019explore} and Liu~\etal~\cite{liu2019guided} are two novel variants of this line. The critical issue in diffusion is how to calculate the affinity matrix. Some works~\cite{qin2011hello,iscen2017efficient} use the reciprocal neighborhood relations to refine the search results, and other~\cite{philbin2008object,chang2019explore} use the inliers number of SP to re-weight the similarity. However, these methods compress the relations between two images as a single scale. As a result, these prior methods lost the spatial matching detail. Although the propagation methods are elegant, they have difficulties recovering good relations to guide the diffusion direction. 

We propose a novel and straightforward hypergraph model to utilize the relations among database images better. It connects the corresponding local features among images using hyperedges; in a sequence of similar images, it detects the image from which the following images are no longer relevant to the query. This approach significantly enhances both image and pixel retrieval performance.

\subsection{Local descriptors and spatial verification}

Local descriptors are the representations of an image's important patches. Some popular local descriptors are RootSIFT~\cite{arandjelovic2012three}, deep local feature (DELF)~\cite{Noh_2017_ICCV} and deep local and global feature (DELG)~\cite{cao2020unifying}.

Local descriptors' location information is usually used in the spatial verification (SP)~\cite{philbin2007object} stage. A standard SP pipeline is first using a KD-tree~\cite{philbin2007object} to find the nearest neighbors between local descriptors in two images. Then it uses the RANSAC algorithm~\cite{fischler1981random} to estimate the homography matrix~\cite{szeliski2010computer} and the number of inlier correspondences between two images.  Existing image search engines commonly apply SP on the shortlist, say the top 100 images ranked by a global descriptor.

SP is known to be crucial for verifying true positive images~\cite{radenovic2018revisiting}. Performing SP before diffusion improves the final retrieval performance~\cite{radenovic2018revisiting, chang2019explore}. However, it is also the slowest process~\cite{arandjelovic2012three,philbin2008object} in image search. In this paper, we put the heavy SP in an offline setting and propose the hypergraph diffusion method to propagate the spatial matching information online. In addition, we also introduce the community selection technique to reduce SP's computing overhead. Community selection frames the diffusion initialization task as selecting a community for the given query rather than verifying the top-ranked images individually. The experiment shows that our hypergraph diffusion and community selection reduce SP overhead without hurting accuracy. 

\section{Overview}
\label{sec:method}

We first discuss issues of query expansions and diffusion process of ordinary graphs in Sec.\ref{sec: generalized function} and our novel hypergraph based propagation (Sec.~\ref{sec_ext}) and initialization scheme utilizing the concept of community  (Sec.~\ref{sec_new_query}) for achieving accurate diffusion process.

\begin{figure*}
    \centering
    \includegraphics[width=0.95\textwidth]{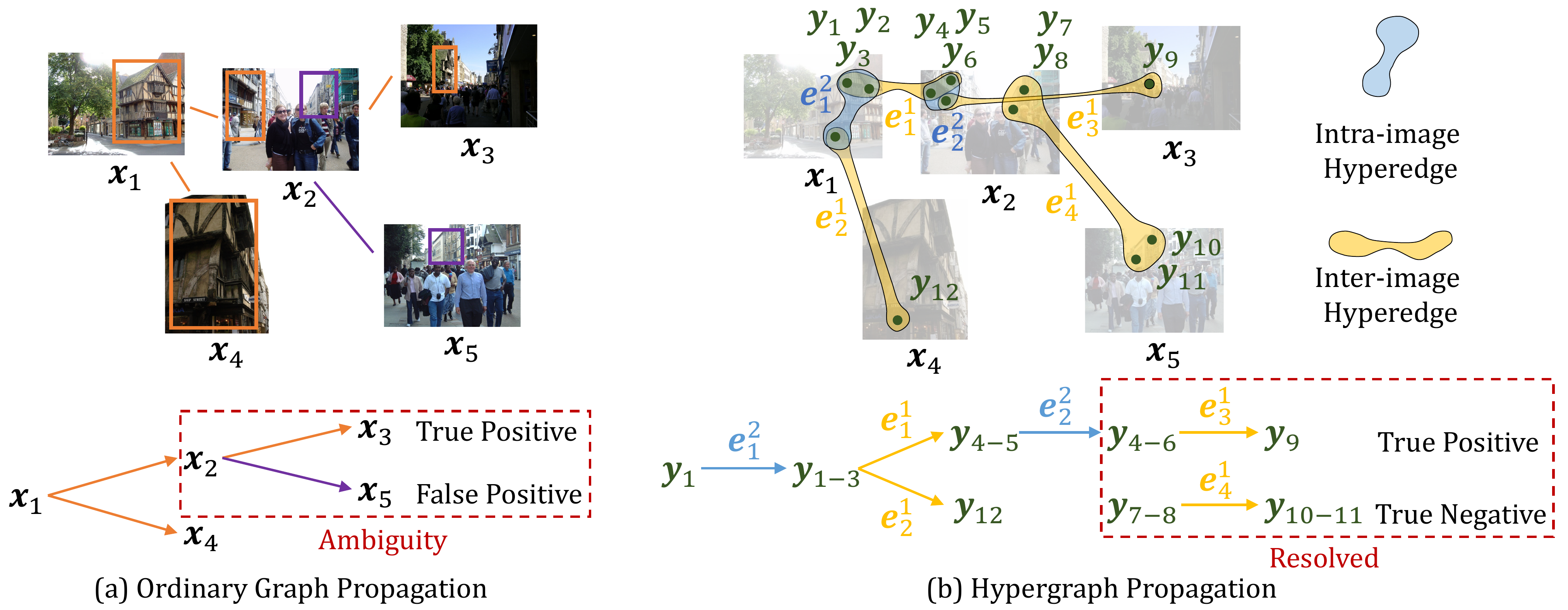}
    \caption{ a) shows a part of an ordinary graph with scalar-weighted, i.e., similarity, edges. Orange frames are the common visible regions among images $\textbf{x}_1$, $\textbf{x}_2$, $\textbf{x}_3$, and $\textbf{x}_4$. Purple frames are the common visible regions between images $\textbf{x}_2$ and $\textbf{x}_5$. $\textbf{x}_3$ and $\textbf{x}_5$ are close neighbors to image $\textbf{x}_2$. While $\textbf{x}_3$ is 
   related to  $\textbf{x}_1$ by sharing the orange frame, $\textbf{x}_5$ is not. Utilizing scalar-weighted edges cannot propagate the query in the ordinary graph without this ambiguity issue. b) shows the corresponding hypergraph of a).
  Inter-image hyperedges $\textbf{e}^1_s$ are shown in yellow, intra-image hyperedges $\textbf{e}^2_k$ are in blue, and local features $\textbf{y}_n$ are in green. A hypergraph path connects local features from $\textbf{y}_1$ to $\textbf{y}_{9}$ in $\textbf{x}_1$ and $\textbf{x}_3$, but no path connects local features in $\textbf{x}_1$ and $\textbf{x}_5$. A large version of this figure is in the appendix.
  }
    \label{fig:graph}
\end{figure*}

\subsection{Propagating in ordinary graph}  
\label{sec: generalized function}

We can use a graph $\textbf{\textit{G}}=(\textbf{\textit{X}}, \textbf{\textit{F}})$ to represent the database images. $\textbf{\textit{X}}:= \left\{ \textbf{x}_1,...,\textbf{x}_m \right\}$ is the node set, where item $\textbf{x}_i$ could be either an image or an image region depending on a chosen retrieval method; $\textbf{\textit{F}}$ is the set of edges whose scalar weight represents the similarity between two nodes connected by each edge.   
Utilizing information in this ordinary graph is an important strategy to improve retrieval results, and, thus, widely used in various methods, such as query expansion~\cite{chum2007total,chum2011total,radenovic2018fine,gordo2020attention} and diffusion~\cite{iscen2017efficient,zhou2004ranking}. Both diffusion and query expansion can be represented by the following function:

\begin{equation}
    \textbf{f}^*=I ( \textbf{f}^{0} ),
    \label{eq:diffusion}
\end{equation}
where $I$ represents the post-processing process, $f_u \in \mathbb{R}$ is the ranking score for $\textbf{x}_u$, $\textbf{f}^0$ is the initial ranking vector with $f_u^0 = 1$ if $\textbf{x}_u$ is a query and $f_u^0=0$ otherwise, and  $\textbf{f}^*$ is the ranking vector after query expansion or diffusion. Diffusion methods diffuse the label of the query node to its neighbors along the graph edges until reaching a stationary state. The stationary state is found by repeating the iteration function: $\textbf{f}^{t+1}=\alpha \textbf{O} \textbf{f}^t + (1-\alpha)\textbf{f}^0$, where $\textbf{O}$ is the normalized affinity matrix that describes the weights of graph edges, $\alpha$ is a jumping probability. Query expansion gathers useful information in the query node's nearest neighbors to update the ranking list. 


Diffusion and query expansion have different advantages and disadvantages. Diffusion can efficiently propagate information to the whole graph through a compact formulation. However, the propagation is conducted through the affinity matrix $\textbf{O}$~\cite{zhou2004ranking,iscen2017efficient}, whose scalar value cannot represent accurate spatial information, resulting in low precision. Take Figure~\ref{fig:graph} as an example, where images of $\textbf{x}_1$ and $\textbf{x}_5$ are incorrectly connected through image $\textbf{x}_2$ in the ordinary graph with scalar edge weights; note that they do not share common regions.
This leads to false positives in the diffusion process. We denote this as the \textbf{ambiguity problem of propagation}. 
In contrast, query expansion uses sophisticated approaches, such as transformer~\cite{gordo2020attention} and spatial verification~\cite{chum2007total,arandjelovic2012three}, to guarantee that only shared information between query and its nearest neighbors are used to update the ranking list. This approach, however, only explores the neighborhood of very similar images due to the computational overhead~\cite{radenovic2018revisiting,arandjelovic2012three,chum2007total}, resulting in a low recall. 

This paper proposes novel hypergraph propagation and community selection, settling the ambiguity problem separately in the propagation and initialization stages. We describe these two techniques next.

\subsection{Extension to hypergraph}
\label{sec_ext}


One meaningful direction to overcome the drawbacks of diffusion and query expansion is to handle the ambiguity problem of graph propagation by explicitly resolving the propagation in a more informative space. The local features contain informative spatial information that can help to distinguish the ambiguity through geometric verification~\cite{philbin2007object}. However, considering the large scale of the image database, it is impractical to do the geometric verification on all the pairs. In practice, image search engines usually build the kNN graph instead of the complete graph, which means every image only connects with its top k similar images~\cite{iscen2017efficient}. To utilize the spatial information on the kNN graph, we propose a novel hypergraph propagation model to efficiently propagate spatial information through hyperedges, connecting an arbitrary number of nodes. This technique is appropriate for the real-world situation where the same object in different images has a different number of local features. In the example of Figure~\ref{fig:graph}, image $\textbf{x}_1$ can connect image $\textbf{x}_3$ by matching the local features with those of image $\textbf{x}_2$, but will not connect image $\textbf{x}_5$. 

When a query is given, we define a hypergraph $\textbf{\textit{H}}=(\textbf{\textit{Y}}, \textbf{\textit{E}})$ on top of the ordinary image graph, where 
its node set $\textbf{\textit{Y}}= \left\{ \textbf{y}_1,...,\textbf{y}_n \right\}$ contains all the local features extracted, and the edge set $\textbf{\textit{E}}$ consists of hyperedges, which are used to connect the local features related to the query object in different images.
Our model then propagates the label of local features of the query to their neighbors and uses an aggregation function to find the final ranking list of images, as follows:
\begin{equation}
    \textbf{l}^*=I_h ( \textbf{l}^0 ), \textbf{f}^*=Agg ( \textbf{l}^*) ,
    \label{eq:hyperdiffusion}
\end{equation} 
where $\textbf{l}^0$ and $\textbf{l}^*$ are the initial and final ranking lists of local features respectively, $\textbf{f}^*$ is the ranking list of images, $I_h$ is the propagation function in a hypergraph, and $Agg$ is the aggregation function that converts ranking scores of local features to ranking scores of images.
We will discuss the construction of hypergraph in Section~\ref{sec:hypergraph} and the propagation process in Section~\ref{sec:diffusion}.

\subsection{Handling new queries: community selection}
\label{sec_new_query}


Diffusion methods~\cite{donoser2013diffusion,zhou2004learning} usually consider the query point to be contained in the dataset. The standard approach to handle a new query is representing the query using its top-ranked images in the initial search to start the propagation~\cite{iscen2017efficient}. However, it is hard to guarantee that these global descriptor-based nearest neighbors contain the same object as the query. We call it the \textbf{ambiguity problem of initialization}.  Recent studies~\cite{radenovic2018revisiting,chang2019explore} perform online spatial verification on 100 nearest neighbors of the query and delete the wrong neighbors to improve the quality of diffusion. Unfortunately, conducting spatial verification for 100 images in query time is a heavy computational load for a search engine. 

 
We find that the structure of the image graph~\cite{philbin2008object,arandjelovic2012three} contains helpful information to measure the accuracy of the nearest neighbors. Following the term in graph theory, we call tightly connected groups communities, which are sets of nodes with many connections inside and few to outside. We observe that the items containing the same objects have a high probability of composing a community and having short geodesic distances.
Based on this observation, we frame the diffusion initialization task as selecting a community for the given query instead of using spatial verification to delete the wrong neighbors. This graph-based approach
improves both accuracy and speed. We will give the detail of community selection in Section~\ref{sec:community selection}

\section{Methods}
\label{sec:hypergraph diffusion}

\subsection{Construction}
\label{sec:hypergraph}

Assume that we have an image dataset $\textbf{\textit{X}}= \left\{ \textbf{x}_1,...,\textbf{x}_m \right\}$, where $\textbf{x}_u$ is an image, and its corresponding K-nearest neighbor (Knn) graph~\cite{zhou2004ranking,dong2011efficient} $\textbf{\textit{G}}=(\textbf{\textit{X}},\textbf{\textit{F}} )$, where $\textbf{\textit{F}}$ is the ordinary edge sets. For hypergraph propagation, we define a hypergraph $\textbf{\textit{H}}=(\textbf{\textit{Y}}, \textbf{\textit{E}})$, where
$\textbf{\textit{Y}}= \left\{ \textbf{y}_1,...,\textbf{y}_n \right\}$ is the set of vertices and $\textbf{y}_i$ represents a local feature in an image.
$\textbf{\textit{E}} =\textbf{\textit{E}}^1 \cup \textbf{\textit{E}}^2$ is the 
union of the inter-image hyperedges set $\textbf{\textit{E}}^1=\left\{\textbf{e}^1_1,\textbf{e}^1_2,...\textbf{e}^1_s \right\}$ and the intra-image hyperedges set $\textbf{\textit{E}}^2=\left\{\textbf{e}^2_1,\textbf{e}^2_2,...\textbf{e}^2_k\right\}$. An inter-image hyperedge connects local features in two images; we also say it connects two images in this paper.  An intra-image hyperedge connects local features in the same image. These hyperedges are shown by the orange and blue groups in Figure~\ref{fig:graph}-b).
The hypergraph has two incidence matrixes $\textbf{S}^1=(s^1_{ia})$ and $\textbf{S}^2=(s^2_{ia})$, where $s^\alpha_{ia}$ is 1 if $\textbf{y}_i \in \textbf{e}^\alpha_a$ and 0 otherwise. We build the ordinary Knn offline and build the hypergraph on top of the Knn graph online. We call the local features in the hypergraph activated local features. We now describe how to construct the hypergraph.

In order to solve the ambiguity problem of graph propagation, hypergraph propagation propagates only among the spatially matched local features of images. This paper uses RANSAC~\cite{fischler1981random}, the most common features matching method in the spatial verification stage of image search~\cite{radenovic2018revisiting,Noh_2017_ICCV,cao2020unifying}, to pre-compute the homography matrix~\cite{fischler1981random} between two images offline. We then use these pre-computed homography matrixes to verify if two features in two images are matched when building the hyperedges at query time. 

We assume the query image can be represented by local features in $\textbf{\textit{Y}}$ and will discuss how to handle a new query in Section~\ref{sec:community selection}. Because a propagation process of a query is progressive, we expand hyperedges on-demand.
To judge if an image is related to the query in a sequence of similar images, we only need to see if it has the activated local features. An inter-image hyperedge is then defined as follows: 

\begin{definition}[Inter-image hyperedge]
\label{def:inter}
For two images $\textbf{x}_u$ and $\textbf{x}_v$, let $\textbf{\textit{V}}_u \subset \textbf{Y}$ and $\textbf{\textit{I}}_v \subset \textbf{Y}$ be the spatially matched local feature sets between $\textbf{x}_u$ and $\textbf{x}_v$. The inter-image hyperedge from features in $\textbf{x}_u$ to $\textbf{x}_v$ is:
$\textbf{e}^1= \left(  In= \textit{V}_u, OUT=\textit{I}_v \right)$.

\end{definition}


If the matching inliers between two images are less than a threshold, they could be directly verified as irrelevant with each other. As the target of hypergraph is to search the related images for the query, we do not build the hyperedge for them. We set the threshold as 20 in this paper, which is the standard threshold in the spatial verification stage of the image search systems~\cite{philbin2008object,chum2009large}. Note that even if an image has more than 20 matching inliers with another one, its activated local features may not be matched.


Similar to object detection~\cite{ren2015faster}, we use bounding box to describe the object region. In an image, local features depicting an object locate in the object's bounding box. As a homography does not guarantee to find all the matched features between two images, the inter-image hyperedges do not connect all the features in an object's bounding box. Unfortunately, the unmatched related features are also important for propagation. As shown in Figure~\ref{fig:graph}-b), features $\textbf{y}_1$, $\textbf{y}_2$ and $\textbf{y}_3$ are in the same region and describing the same building, but only $\textbf{y}_1$ and $\textbf{y}_3$ are connected within image $\textbf{x}_2$ using an inter-image hyperedge to image $\textbf{x}_1$. Without an intra-image hyperedge connecting $\textbf{y}_1$, $\textbf{y}_2$ and $\textbf{y}_3$, we cannot propagate information from images $\textbf{x}_1$ to $\textbf{x}_3$. 
To recover the unmatched yet related local features, we define the intra-image hypergraph as follows:
\begin{definition}[Intra-image hyperedge]
\label{def:inner}
Intra-image hyperedges are built after inter-image hyperedges. Given a feature set in an image $\textbf{x}_u$,
let $\textbf{\textit{V}}_u \subset \textbf{\textit{Y}}$ be the local features that are connected by an inter-image hyperedge in $\textbf{x}_u$, and $\textbf{\textit{I}}_u \subset \textbf{\textit{Y}}$ be the set of all the local features of $\textbf{x}_u$. Let the smallest rectangle covering all the features in $\textbf{\textit{V}}_u$ be the bounding box $b$. The intra-image hyperedge in $\textbf{x}_u$ is defined as $\textbf{e}^2 =$ ( $IN= \textbf{\textit{V}}_u, OUT=\{ \textbf{y}_j \in \textbf{\textit{I}}_u; \textbf{y}_j \mbox{ exists in } b \} ) $.

\end{definition}

Due to the high computing cost, it is impractical to perform the propagation on a large-scale dataset consisting of millions of images. Inspired by the idea of truncation~\cite{iscen2017efficient} in diffusion, we build inter-image hyperedges only among the hop-$N$ neighbors~\footnote{Node A is in B's hop-$N$ neighbors if A can be connected with B using not more than $N$ different hyperedges.} of a given query image. Specifically, we build the inter-image hyperedges from image $\textbf{x}_u$ to $\textbf{x}_v$ only if both of them are in the hop-$N$ neighbors of the query.


\subsection{Propagation}
\label{sec:diffusion}

After building the hypergraph, we use a diffusion model in the hypergraph to calculate the final ranking scores of retrieved images.
Specifically, we define a diffusion matrix $\textbf{P}=(p_{ij})$ and set the diffusion weight between $\textbf{y}_i \in \textbf{x}_u$ 
and $\textbf{y}_j \in \textbf{x}_v$ as $p_{ij} = d(\textbf{x}_u, \textbf{x}_v)  \max_{akb} (s^1_{ia}  s^1_{ka}  s^2_{kb}  s^2_{jb} ) $, where $d(\textbf{x}_u, \textbf{x}_v)$ is the Euclidean distance between the global features of $\textbf{x}_u$ and $\textbf{x}_v$, and $\max_{akb} (s^1_{ia}  s^1_{ka}  s^2_{kb}  s^2_{jb} )$ records whether $\textbf{y}_i$ and $\textbf{y}_j$ are connected in the hypergraph. More specifically,  $ s^1_{ia}  s^1_{ka}  s^2_{kb}  s^2_{jb}$ indicates that  $\textbf{y}_i$ and $\textbf{y}_j$ can be connected through a intermediate node $\textbf{y}_k$; $s^1_{ia}  s^1_{ka} $ shows $\textbf{y}_i$ and $\textbf{y}_k$ are connected through inter-image hyperedge $\textbf{e}_a^1$, and $ s^2_{kb}  s^2_{jb} $ shows $\textbf{y}_k$ and $\textbf{y}_j$ are connected through intra-image hyperedge $\textbf{e}_b^2$. If $k = i$ or $k=j$, $\textbf{y}_i$ and $\textbf{y}_j$ are directly connected. 
The diffusion matrix is normalized using the degree matrix $\textbf{D}=diag(\textbf{P}\textbf{1}_n) $ as $\textbf{P}'=\textbf{P}\textbf{D}$. For aggregation function, $a_{iu} \in \textbf{A}$ is set as $\frac{1}{b_u}$ if $\textbf{y}_i \in \textbf{x}_u$, and 0 otherwise, where $b_u$ is the number of activated local features 
in $\textbf{x}_u$. Then the diffusion function in Equation~\ref{eq:hyperdiffusion} can be written as:
\begin{equation}
    \begin{aligned}
        \textbf{Y}^* &= I_h(\textbf{Y}^0)=\textbf{Y}^0+\textbf{Y}^0 \textbf{P}'+\textbf{Y}^0 \textbf{P}'^2+ \cdots + \textbf{Y}^0 \textbf{P}'^N, \\
        \textbf{f}^* &= Agg(\textbf{Y}^*)=\textbf{Y}^*\textbf{A},
    \end{aligned}
    \label{eq:heuristic}
\end{equation}
where $\textbf{Y}^0$ is the initial score vector for local features. In $\textbf{Y}^0$, the scores of local features in the query image are set as 1, while others are 0. The iteration version for Equation~\ref{eq:heuristic} is $\textbf{Y}^{t}= \textbf{Y}^{t-1}\textbf{P}'+\textbf{Y}^0$. When $t=N$ and $t$ is large, $\textbf{Y}^t $ is converged to $\textbf{Y}^*=\textbf{Y}^0+\textbf{Y}^0 \textbf{P}'+\textbf{Y}^0 \textbf{P}'^2+ \cdots + \textbf{Y}^0 \textbf{P}'^N = (\textbf{I}-\textbf{P}')^{-1}\textbf{Y}^0$; this convergence is guaranteed because $ \sum_i p_{ij} <1 $ based on its definition and $\textbf{Y}^0+\textbf{Y}^0 \textbf{P}'+\textbf{Y}^0 \textbf{P}'^2+ \cdots + \textbf{Y}^0 \textbf{P}'^N$ is the power series representation of $(\textbf{I}-\textbf{P}')^{-1}\textbf{Y}^0$. 
Intuitively, this diffusion function
gives high scores to the images with shorter hypergraph geodesic distances to the query while considering similarities of global features. 

This diffusion method can simultaneously achieve pixel retrieval. After diffusion, we treat all the local features with non-zero scores as the related local features of the target object. For each retrieved image, we draw the related local features' bounding box or polygon as the target object region.


\subsection{Community selection for new query}
\label{sec:community selection}

The hypergraph propagation described in Sec.~\ref{sec:hypergraph diffusion} assumes the query is in the database, which may not be true in a general retrieval scenario. Now we consider how to find the good starting nodes for a new query. 

\begin{figure}[]
    \centering
    \includegraphics[width=0.45\textwidth]{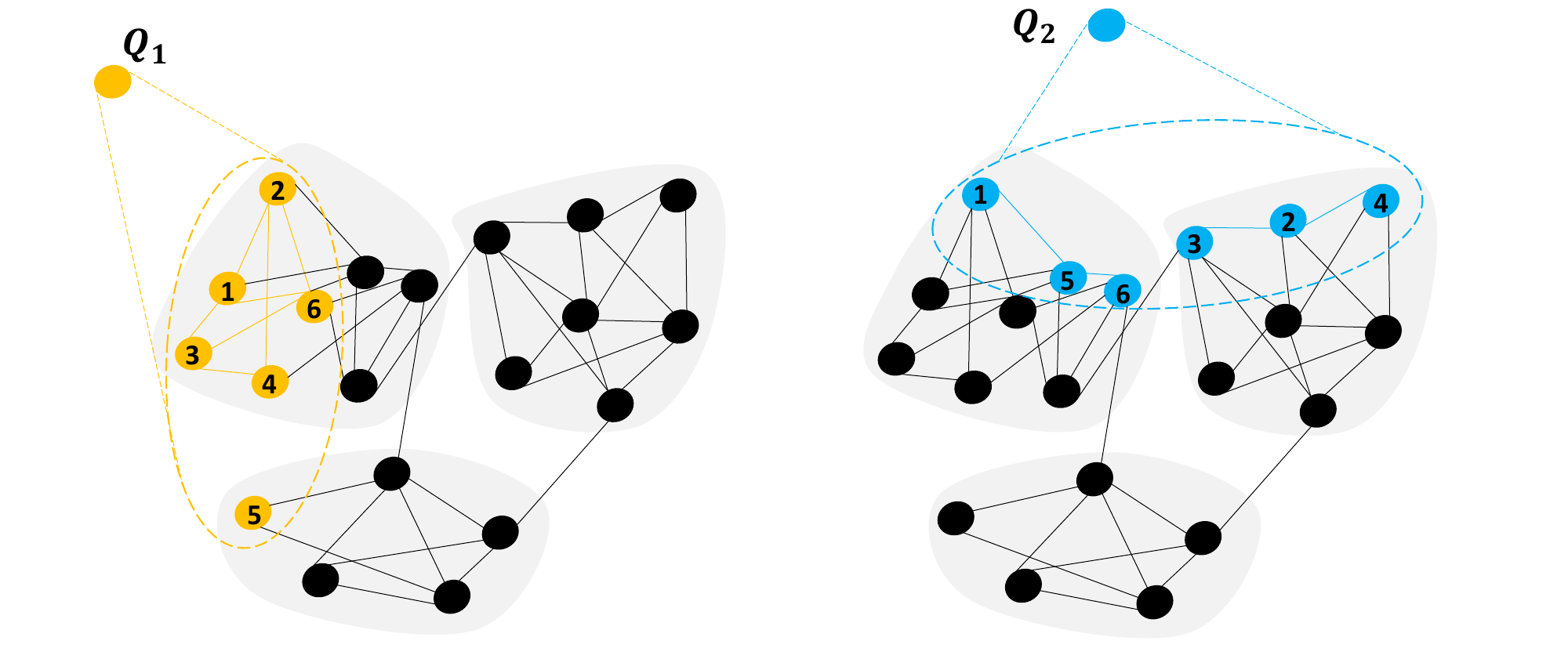}
    \caption{Two queries on an image graph with three communities. The numbers on the nodes are their rankings in the initial search. The uncertainty of the initial search result of Q1 is lower than that of Q2 as most retrieved items of Q1 distribute in the same community.}
    \label{fig:community}
\end{figure}

As discussed in Section~\ref{sec:method}, 
we observe that the items containing the same object usually belong to the same community in $G$. By checking whether the top S items of the initial search are in the same community, we can evaluate the quality of the initial search without query time spatial verification. As shown in Figure~\ref{fig:community}, the quality of the global-feature-based search of Q1 is better than Q2 because its top-ranked items are in the same community. 



Specifically, after the initial search, we use the top $S$ images and their connected edges in $G$ to build a subgraph
$G_{s}$. $G_{s}$ consists of one or more components,
and each component represents a community in $G$. We call the top-1 image of initial search the dominant image, its component in $G_s$ the dominant component, and its community in $G$ the dominant community. 
We call other components in $G_{s}$ as the opposition components. Intuitively if all the top $S$ images in $G_s$ constitute a single component, all of them may contain the same object with the query image. Otherwise, the uncertainty will increase if the number and size of the opposition components increase. This observation also agrees with our experimental results. Inspired by Shannon entropy, we define the uncertainty index, $U$, of the initial search as:
 \begin{equation}
     U=-\sum_{i=1}^{n_G} P\left(C_{i}\right)\log P \left( C_{i}\right),
 \end{equation}
 where $n_G$ is the number of component in $G_{s}$, $C_{i}$ is a component and $P\left(C_{i}\right)$ is the proportion of $C_{i}$ in $G_{s}$.

 Community selection is the pre-stage of hypergraph propagation. The overall retrieval process with community selection is to perform the initial search firstly and then calculate the uncertainty of the initial search result.  If the uncertainty index is low, say $U$ is lower than 1, hypergraph propagation uses images in the dominant community to represent the query image and starts the propagation process from them. If the uncertainty index is high, the search engine does the spatial verification on the top $T$ images of the initial search to find a new dominant community for the query. In this way, we significantly reduce the number of spatial verification while keeping very high accuracy. 

 \section{Experiment}
 \label{sec:experi}

We first introduce the benchmarks and implementation detail in Section~\ref{subsec:benchmarks} and~\ref{subsec:implementation}, respectively. We then compare the accuracy of our hypergraph diffusion and the SOTA image-level and pixel-level methods in Section~\ref{subsec:image-level result} and~\ref{subsec:pixel result}. We show the effectiveness of community selection in Section~\ref{subsec:community selection result}. Finally, we report and discuss the time and memory costs in Section~\ref{subsec:time-memory}. 

\subsection{Test datasets and evaluation protocals}
\label{subsec:benchmarks}

For image-level retrieval, we evaluate our methods on two well-known landmark retrieval benchmarks: revisited Oxford (ROxf) and revisited Paris (RPar)~\cite{radenovic2018revisiting}. There are 4,993 (6,322) database images in the ROxf (RPar) dataset, and a different query set for each dataset, both with 70 images. We also report the result with the R1M distractor set, which contains 1M images. The accuracy is measured using mean average precision (mAP).

For the pixel-level retrieval, we evaluate our methods on pixel-retrieval benchmarks: pixel revisited Oxford (PROxford) and pixel revisited Paris (PRParis)~\cite{an2023towards}. These two datasets are designed based on the famous ROxford and RParis~\cite{radenovic2018revisiting}. They have the same query, database, and distractor images as ROxford and RParis, and they provide pixel-level annotation. The accuracy is measured using mAP@50:5:95. Each ground truth image in the ranking list is treated as a true-positive only if its detection Intersection over
Union (IoU) is larger than a threshold n; the threshold n is set from 0.5 to 0.95, with step 0.05. As~\cite{an2023towards}, we report the mean of IoU (mIoU) for all the methods and report the mAP@50:5:95 for some representative methods.

 \subsection{Implementation}
 \label{subsec:implementation}

We use the DELG model~\cite{cao2020unifying} trained on GLDv2~\cite{weyand2020google} to extract both global and local features for our Hypergraph Diffusion (HD) and all the baseline methods unless a specific explanation is given. We find the nearest neighbor number $K$ does not influence the final result of HD a lot, and we report the performance of setting $K$ as 200. 

For the image-level comparison, we evaluate the performance of HD and baseline methods under the ordinary retrieval pipeline: performing the initial search, followed by conducting reranking methods like query expansion and diffusion. We reorder the initial ranking list using $\textbf{f}^*$ after the HD as the final results. 
We do not use community selection and always start the HD from the initial dominant community for all the queries to make the comparison fair. 

For the pixel-level comparison, we test all the methods using the same image-level ranking result, the image ranking result obtained by our HD, for a fair comparison. HD directly produces the pixel-level result when reranking the image-level result without having any extra time-consumption.
For the SOTA baseline methods, we use them to calculate for each database and query pair to get the pixel-level result. The accuracy is reported in Section~\ref{subsec:pixel result}, and the time and memory cost is reported in Section~\ref{subsec:time-memory}.


For community selection, we set $S$, the number of images for calculating the uncertainty, as 20. When testing the performance of combining community selection and hypergraph propagation, we set the uncertainty threshold as 1, which we find to be a good balance between the accuracy and computation overhead. For the query with an uncertainty index higher than the threshold, we do the spatial verification for the top 100 items using SIFT features~\cite{lowe1999object}. Once we find an item with more than 20 inliers, we treat its community as the new dominant community, stop the spatial verification, and apply hypergraph propagation in the new dominant community. 

We conduct the offline process on 6 Intel(R) Core(TM) i9-9900K CPUs @ 3.60GHz and 896GB of RAM and implement the online hypergraph propagation and community selection on one CPU and 50GB of memory.
The code of this work is publicly available on \url{https://sgvr.kaist.ac.kr/~guoyuan/hypergraph_propagation/}

\begin{table*}[]
    \caption{Results (\% mAP) on the ROxf/RPar datasets and their large-scale versions ROxf+1M/RPar+1M, with both Medium and Hard evaluation protocols. \textbf{Bold} number indicates the best performance, and \underline{underline} indicate the second and third one.}
    \label{tab:accuracy}
    \centering
    \small
    
    \setlength{\tabcolsep}{0pt}
    \newcolumntype{C}[1]{>{\centering\arraybackslash}p{#1}}
    \begin{tabular*}{\textwidth}{|C{0.273\textwidth}|C{0.09\textwidth}|C{0.09\textwidth}|C{0.09\textwidth}|C{0.09\textwidth}|C{0.09\textwidth}|C{0.09\textwidth}|C{0.09\textwidth}|C{0.09\textwidth}|}
      \hline
    & \multicolumn{2}{c|}{ROxf} & \multicolumn{2}{c|}{ROxf+R1M} & \multicolumn{2}{c|}{RPar} & \multicolumn{2}{c|}{RPar+R1M}  \\ \hline
         Method & M & H & M & H & M & H & M & H  \\ \hline
         \multicolumn{9}{|c|}{Using standard DELG features} \\ \hline
        Global search~\cite{cao2020unifying} & 76.3 & 55.6 &63.7 &37.5 & 86.6 & 72.4 &70.6 &46.9  \\
        Spatial verification~\cite{cao2020unifying} & 81.2 & 64.0 &69.1 &47.5 & 87.2 & 72.8 &71.5 &48.7  \\
        \hline
        Average QE~\cite{chum2007total} & 77.2 & 57.1 &68.5 &43.0 & 87.6 & 74.3 &75.4 &54.8  \\
        Average QE with decay~\cite{gordo2017end} & 78.4 & 58.0 &70.4 &44.7 & 88.2 & 75.3 &76.2 &56.0  \\
        $\alpha$ QE~\cite{radenovic2018fine} & 65.2 & 43.2 & 57.0 & 30.2 & 91.0 & 81.2 & 81.0 & 64.1 \\
        Diffusion~\cite{iscen2017efficient} & 81.0 & 59.3 &63.9 &38.7 & 91.4 & 82.7 &80.0 &64.9  \\
        Hypergraph Diffusion (Ours) & \textbf{85.7} & \textbf{70.3} &\textbf{78.0} &\textbf{60.0} & \textbf{92.6} & \textbf{83.3} &\textbf{86.6} &\underline{72.7}  \\
        \hline
        \multicolumn{9}{|c|}{Advanced reranking methods} \\      \hline
        LAttQE~\cite{gordo2020attention}&73.4 &49.6&58.3 &31.0 &86.3  &70.6&67.3 &42.4  \\
        LAttDBA+LAttQE~\cite{gordo2020attention}&74.0&54.1&60.0 &36.3 &87.8  &74.1&70.5 &48.3  \\
        SAA~\cite{ouyang2021contextual}&78.2&59.1&61.5 &38.2 &88.2  &75.3&71.6 &51.0  \\
        GSS~\cite{liu2019guided}+SAA~\cite{ouyang2021contextual}&79.3&62.2&62.1 &42.3 &90.7  &80.0&85.1 &\underline{70.3}  \\
        Diffusion~\cite{iscen2017efficient}+SAA~\cite{ouyang2021contextual}&76.3&57.8&66.2 &42.4 &90.2  &81.2&\underline{86.3} &\textbf{75.4}  \\
        \hline
        \multicolumn{9}{|c|}{Advanced deep learning models fine-tuned using SfM} \\      \hline
        HOW (R50)~\cite{tolias2020learning}&78.3 &55.8&63.6 &36.8 &80.1  &60.1 &58.4 &30.7  \\
         FIRe (R50)~\cite{weinzaepfel2022learning}&81.8  &61.2&66.5 &40.1 &85.3  &70.0 &67.6 &42.9  \\
         
        \hline
        \multicolumn{9}{|c|}{Advanced deep learning models fine-tuned using GLD} \\     \hline
         
        D2R-DELF-ASMK (R50)~\cite{teichmann2019detect}&76.0 &52.4&64.0 &38.1 &80.2  &58.6 &59.7 &29.4  \\
        
        Token (R101)~\cite{wu2022learning}&77.8 &60.1&66.7 &43.1 &87.9  &74.7 &75.2 &53.2  \\
        DOLG (R101)~\cite{yang2021dolg} &78.4 &58.6&75.5 &52.4 &88.5  &75.4 &78.3 &61.9  \\
        SOLAR (R101)~\cite{ng2020solar} &81.6 &63.3&71.8 &45.3 &88.2  &75.2 &72.9 &51.3  \\
        SpCa-cro (R50)~\cite{zhang2023learning}&79.9 &59.3&72.8 &49.3 &87.4  &73.1 &78.0 &58.3  \\
        SpCa-cat (R50)~\cite{zhang2023learning} &81.6 &61.2&73.2 &48.8 &88.6  &76.2 &78.2 &60.9  \\
        SpCa-cro (R101)~\cite{zhang2023learning}&82.7 &65.6&\underline{77.8} &\underline{53.4} &90.2  &79.3 &79.1 &65.8  \\
        SpCa-cat (R101)~\cite{zhang2023learning}&83.2 &65.9&\underline{77.8} &\underline{53.3} &90.6  &80.0 &79.5 &65.0  \\
        
        CFCD-3 scales (R50)~\cite{zhu2023coarse}&82.5 &63.6&72.7 &48.5&89.6  &78.1 &78.9 &60.1  \\
        CFCD-3 scales (R101)~\cite{zhu2023coarse}&\underline{84.1} &\underline{67.8}&74.7 &54.1&91.0  &81.2&82.2&65.5  \\
        CFCD-5 scales (R50)~\cite{zhu2023coarse}&82.4 &65.1&73.1 &50.8&\underline{91.6}  &\underline{81.7}&81.6&62.8  \\
        CFCD-5 scales (R101)~\cite{zhu2023coarse}&\underline{85.2} &\underline{70.0}&74.0 &52.8&\underline{91.6}  &\underline{81.8}&82.8&65.8  \\
        \hline

    \end{tabular*}

\end{table*}

\subsection{Comparison to the image retrieval state-of-the-art}
\label{subsec:image-level result}

\begin{figure*}[h]
    \centering
    \includegraphics[width=\textwidth]{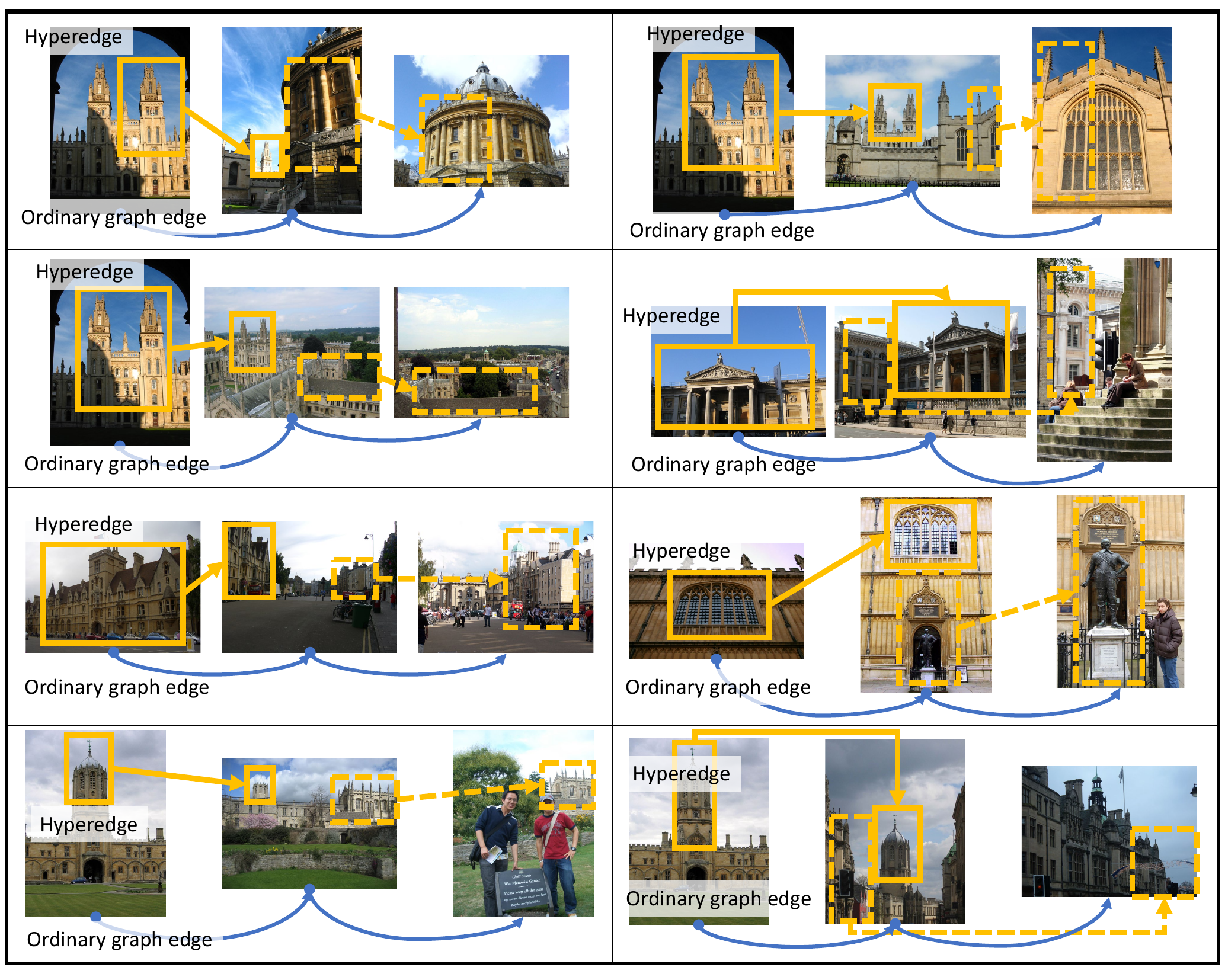}
    \caption{Illustration of hypergraph diffusion mechanism. The orange boxes with arrows represent the hyperedges, and the blue curved arrows are the ordinary graph edges. In each triplet, the first image and the third image are wrongly connected through the second image. While traditional diffusion and QE methods wrongly propagate the similarity score from the first to the third image through ordinary graph edge, our hypergraph diffusion does not diffuse the similarity scores from the first to the third image by solving the spatial ambiguity problem of propagation.  
    }
    \label{fig:HD_vis}
\end{figure*}

\textbf{Baselines and implementation}.
We compare with the result of global search and SPatial verification (SP) using DELG features. For SP, we use the standard RANSAC implemented in the scikit-learn package to predict the affine transform between two images. The initial local feature matching pairs for RANSAC are first going through a ratio test with a threshold of 0.9. We set the maximum RANSAC trials as 1000. This setting is also used by the origin DELG paper~\cite{cao2020unifying}. We tried several different settings during the experiment, such as changing the ratio test threshold and increasing the maximum RANSAC trials, but did not observe performance gain. 

We compare hypergraph propagation against the existing query expansion and diffusion methods: average query expansion~\cite{chum2007total}, average query expansion with decay~\cite{gordo2017end}, $\alpha$ query expansion, and diffusion~\cite{donoser2013diffusion}. We directly implement query expansion methods and use the open \href{https://github.com/ducha-aiki/manifold-diffusion}{code} from Mishkin to reproduce diffusion. Query expansion and diffusion are applied to global DELG features. Although the efficient diffusion approach~\cite{iscen2017efficient} can work for sparse local features like R-MAC (21 features for each image), we find it is not easy to directly apply it to the dense local features like DELG (each image contains up to 1000 features). We additionally compare with some recent advanced learning-based reranking methods that use dense features, which we will introduce next.

This manuscript builds upon our prior work~\cite{an2021hypergraph}.
During the preparation of this manuscript, several advanced methods have been introduced. We benchmark our approach against these new techniques. We directly report the best-reported performance for each method instead of reproducing them. The new reranking methods are Guided Similarity Separation (GSS)\cite{liu2019guided}, Learnable Attention-based Database-side Augmentation (LAttDBA), Query Expansion (LAttQE)\cite{gordo2020attention}, and Self-Attention Aggregation (SAA)\cite{ouyang2021contextual}. We also evaluate against cutting-edge deep learning models, particularly those fine-tuned on \textbf{Structure-from-Motion (SfM)} datasets, namely HOW\cite{tolias2020learning} and Feature Integration-based Retrieval (FIRe)\cite{weinzaepfel2022learning}, as well as those refined using the \textbf{Google Landmark Dataset (GLD)}, including Detection-to-Retrieval (D2R)\cite{teichmann2019detect}, Token-based model (Token)\cite{wu2022learning}, Deep Orthogonal fusion of Local and Global features (DOLG)\cite{yang2021dolg}, Second-Order Loss and Attention for Image Retrieval (SOLAR)\cite{ng2020solar}, Spatial-Context-Aware model (SpCa)\cite{zhang2023learning}, and Coarse-to-Fine Compact Discriminative model (CFCD)~\cite{zhu2023coarse}.


\textbf{Result}.
Table~\ref{tab:accuracy} compares the accuracy of hypergraph diffusion (HD) and other retrieval methods. Among the QE/diffusion methods, HD obtains the best result on ROxford and RParis, both with and without the R1M distractor set. Compared to RParis, ROxford is more challenging as it contains more difficult retrieval cases such as the change of viewpoint~\cite{radenovic2018revisiting}. The performance improvement of our method on ROxford hard cases is impressive; we get the mAP 70.3 and 60 with and without 1M distractors, outperforming the traditional diffusion method 18.5\% and 55\%, respectively. Our results are also better than other techniques in ~\cite{radenovic2018revisiting,chang2019explore}, which integrate additional steps with more computation and memory requirements.  

In Figure~\ref{fig:HD_vis}, we demonstrate the efficacy of hypergraph propagation using actual query examples. The figure utilizes highlighted regions, marked in yellow, to indicate areas where correct matches are identified through hyperedges in our model. The figure presents triplets of images to illustrate a common issue in traditional diffusion and QE methods. In each triplet, the first and third images are often mistakenly linked via the second image, although they do not contain the same object. This linkage exemplifies the famous false-positive problem in image retrieval, as discussed in~\cite{radenovic2018revisiting}. Our hypergraph mechanism addresses this by resolving the spatial ambiguity commonly observed in propagation processes. By diffusing along the hyperedges instead of the ordinary graph edge, our method significantly mitigates the false-positive issue in existing diffusion and QE techniques. 

Ordinary diffusion is better than QE methods on all the datasets except ROxford with R1M distractors. We think 
that
the texture and shape of medieval buildings in ROxford
are common and tend to appear in other buildings around the world. When retrieving these objects with R1M distractors, its ambiguity problem explores more false positive items, leading to a low mAP than QE methods. Our hypergraph propagation keeps the spatial information during propagation, thus avoiding the false positive issue of ordinary diffusion, resulting in high accuracy. 

While preparing this manuscript, several advanced and promising methods have emerged in the field. As shown in Table~\ref{tab:accuracy}, our HD approach maintains a competitive accuracy compared to these recent methods. HD achieves the highest accuracy on the ROxford and RParis benchmarks. The exception is in the challenging scenario of RParis with 1 million distractors under the hard setting, where the diffusion~\cite{iscen2017efficient} combined with SAA~\cite{ouyang2021contextual} outperforms others.  Table~\ref{tab:accuracy} shows that our HD achieves top-tier accuracy among the existing reported performances of models fine-tuned with standard Google Landmarks Dataset (GLD) and Structure-from-Motion (SfM). These results show that our diffusion approach on dense local features using hypergraphs is promising. Note that only SP and our HD can localize the target object among the methods in Table~\ref{tab:accuracy}, which we will introduce in the next section. 



\subsection{Comparison to the pixel retrieval state-of-the-art}
\label{subsec:pixel result}

\begin{table*}[htp]
    \caption{Results of pixel retrieval from ground truth query-index image pairs (\% mean of mIoU) on the PROxf/PRPar datasets with both Medium and Hard evaluation protocols. D and S indicate detection and segmentation results respectively. \textbf{Bold} number indicates the best performance, and \underline{underline} indicates the second one. Our HD achieves both the best image-level (Table~\ref{tab:accuracy} ) and pixel-level retrieval accuracy among the retrieval and localization unified methods. Note that HD is much faster than SP, detection, segmentation, and dense matching methods, as shown in Table~\ref{tab:pixel-map} and Sec.~\ref{subsec:time-memory}. 
    }
    \label{tab:miou}
    \centering
    \newcolumntype{C}[1]{>{\centering\arraybackslash}p{#1}}
    \begin{tabular*}{\textwidth}{|C{0.354\textwidth}|C{0.045\textwidth}|C{0.045\textwidth}|C{0.045\textwidth}|C{0.045\textwidth}|C{0.045\textwidth}|C{0.045\textwidth}|C{0.045\textwidth}|C{0.045\textwidth}|C{0.045\textwidth}|}
    \hline
   \multirow{3}{*}{Method} & \multicolumn{4}{c}{Medium} & \multicolumn{4}{|c|}{Hard}& \\
    \cline{2-10}
     & \multicolumn{2}{c|}{PROxf}  & \multicolumn{2}{c|}{PRPar} & \multicolumn{2}{c|}{PROxf} & \multicolumn{2}{c|}{PRPar} &Average \\
     \cline{2-10}
          & D & S & D & S & D & S & D & S &\\
    \hline
    \multicolumn{10}{|c|}{ Retrieval and localization unified methods}\\
    \hline
        SIFT+SP~\cite{philbin2007object} &26.1  &10.9 &24.2 &9.7 &18.2  &7.3  &19.3 &7.8&15.44 \\
        DELF+SP~\cite{Noh_2017_ICCV} & \underline{43.7} &20.0  &\underline{40.7} &16.7 &\underline{33.2} &13.9 &32.2 &12.4&26.60 \\
        DELG+SP~\cite{cao2020unifying} &\textbf{44.1} &19.7  &40.1 &16.5 &\textbf{34.8} &14.5 &31.2 &11.7& 26.57 \\
        D2R~\cite{teichmann2019detect}+Resnet-50-Faster-RCNN+Mean  &20.2 &-  &29.6 &- &16.7 &-  &27.4 &-&- \\
        D2R~\cite{teichmann2019detect}+Resnet-50-Faster-RCNN+VLAD~\cite{jegou2010aggregating} &25.8  &- &37.5 &- &21.6  &-  &35.5 &-&- \\
        D2R~\cite{teichmann2019detect}+Resnet-50-Faster-RCNN+ASMK~\cite{tolias2016image} &26.3  &-  &38.5 &- &21.6 &- &\underline{35.6} &-&-\\
        D2R~\cite{teichmann2019detect}+Mobilenet-V2-SSD+Mean &19.7 &-  &25.9 & -&20.1 &-  &27.9 &-&- \\
        D2R~\cite{teichmann2019detect}+Mobilenet-V2-SSD+VLAD~\cite{jegou2010aggregating}  &23.1 &-  &33. &- &20.9 &-  &33.6 &-&- \\
        D2R~\cite{teichmann2019detect}+Mobilenet-V2-SSD+ASMK~\cite{tolias2016image}   &22.4 & - &34.0 &- &20.8 &-  &33.1 &-&- \\
        HD (ours)&29.7 &24.7  &\textbf{49.2}  &38.2  &22.6  &19.4  &\textbf{39.1}  &31.0 &\underline{31.74} \\
    \hline
    \multicolumn{10}{|c|}{Detection methods}\\
    \hline
    OWL-VIT (LiT)~\cite{minderer2022simple} & 11.4 & - & 18.0 & - & 6.3 & - & 15.0 & -&- \\
    OS2D-v2-trained~\cite{osokin20os2d} & 10.5 &-  &13.7 &- &11.7 &- &14.3 &-&-\\
    OS2D-v1~\cite{osokin20os2d}  &7.0 &-  &8.5 &- & 8.7&-  &9.2 &-&- \\
    OS2D-v2-init~\cite{osokin20os2d}   &13.6 &-  &15.4 &- &14.0 &-  &15.1 & -&-\\
\hline
    \multicolumn{10}{|c|}{Segmentation methods}\\
    \hline
    SSP (COCO) + ResNet50~\cite{fan2022ssp} & 19.2 & \underline{34.5} & 31.1 & \underline{48.7} & 15.1 & \underline{25.3} & 29.8 & \textbf{41.7}&30.68 \\
     SSP (VOC) + ResNet50~\cite{fan2022ssp}  & 19.7 & 34.3 & 31.4 & \textbf{48.8} & 16.1 & \textbf{26.1} & 30.3 & \underline{40.4}&30.89 \\
     HSNet (COCO) + ResNet50~\cite{min2021hypercorrelation} & 23.4 & 32.8 & 37.4 & 41.9 & 21.0 & 25.7 & 34.7 & 36.5&31.67 \\
     HSNet (VOC) + ResNet50~\cite{min2021hypercorrelation} & 21.0 & 29.8 & 31.4 & 39.7 & 17.1 & 23.2 & 29.7 & 34.9&28.35 \\
     HSNet (FSS) + ResNet50~\cite{min2021hypercorrelation} & 30.5 & \textbf{35.7} & 39.4 & 40.2 & 22.7 & 25.1 & 34.7 & 32.8&\textbf{32.64} \\
     Mining (VOC) + ResNet50~\cite{yang2021mining} & 18.3 & 30.5 & 29.6 & 42.7 & 15.1 & 21.4 & 28.1 & 34.3&27.50 \\
     Mining (VOC) + ResNet101~\cite{yang2021mining} & 18.1 & 28.6 & 29.5 & 40.0 & 14.2 & 20.4 & 28.2 & 34.4&26.68 \\
        \hline
    \multicolumn{10}{|c|}{Dense matching methods}\\
    \hline
    
    GLUNet-Geometric~\cite{truong2020glu} & 18.1 & 13.2 & 22.8 & 15.2 & 7.7 & 4.6 & 13.3 & 7.8&12.84 \\
    PDCNet-Geometric~\cite{truong2021pdc} & 29.1 & 24.0 & 30.7 & 21.9 & 20.4 & 15.7 & 20.6 & 12.6&21.87 \\
    GOCor-GLUNet-Geometric~\cite{truong2020gocor} & 30.4 & 26.0 & 33.4 & 25.6 & 20.8 & 16.0 & 19.8 & 13.3&23.16 \\
    WarpC-GLUNet-Geometric (megadepth)~\cite{truong2021warp} & 31.3 & 25.4 & 36.6 & 27.3 & 21.9 & 15.8& 26.4 & 17.3&25.25 \\
    GLUNet-Semantic~\cite{truong2020glu} & 18.5 & 14.4 & 22.4 & 15.6 & 8.7 & 5.6 & 12.8 & 7.8&13.22 \\
    WarpC-GLUNet-Semantic~\cite{truong2021warp} & 27.5 & 21.4 & 36.8 & 25.7 & 18.5 & 11.9 & 28.3 & 17.6&23.46 \\
    
     \hline

    \end{tabular*}

\end{table*}

\begin{figure*}[h]
    \centering
    \includegraphics[width=\textwidth]{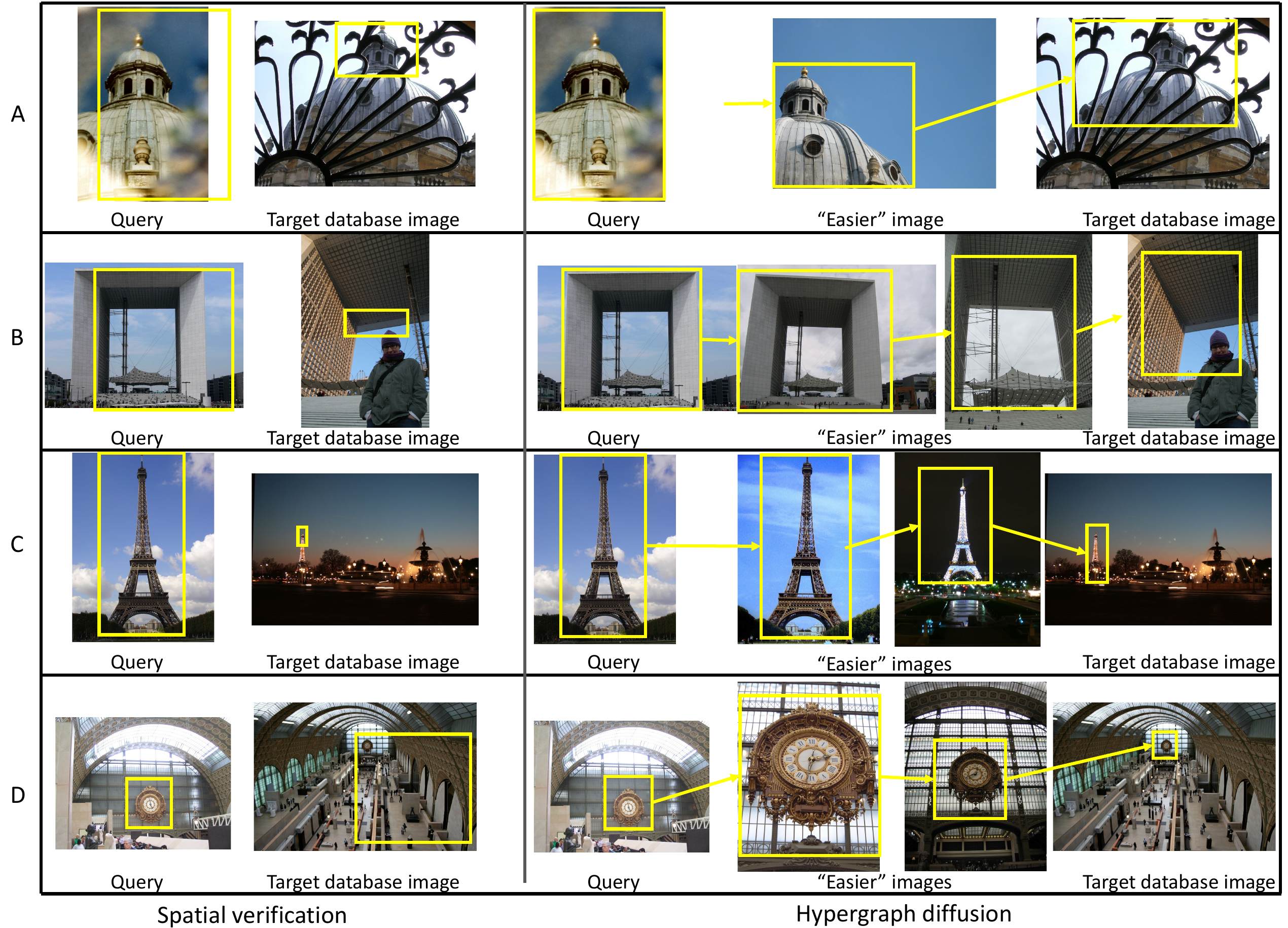}
    \caption{Visual examples of how Hypergraph Diffusion (HD) achieves better pixel retrieval results than direct SPatial verification (SP) using the same DELG features. The yellow boxes in query images are the cropped region given by the benchmark. Yellow boxes in SP are the minimum bounding box of the matching points. Yellow boxes with line arrows indicate the diffusion pathes in HD from the query region to the correspondence region in the target database image. In A, an `easier' database image offers a clear, unoccluded view of the target query landmark, demonstrating improved matching. B highlights how these database images provide viewpoints more aligned with the query image, enhancing object matching. C shows the advantage in scenarios with varying illumination, where `easier' images assist in achieving more accurate matches. Lastly, D reveals that when the target database image has a more complex background than the query image, direct spatial verification can lead to outlier matchings. In contrast, the `easier' database images selected by HD provide additional context, thus facilitating more precise pixel retrieval.  
    }
    \label{fig:pixel_retrieval_vis}
\end{figure*}

\begin{table*}[hbt]
    \caption{Results of pixel retrieval from database (\% mean of mAP@50:5:95) on the PROxf/PRPar datasets and their large-scale versions PROxf+1M/PRPar+1M, with both Medium (M) and Hard (H) evaluation protocols. \textbf{\textcolor{red}{Red}} indicates the best performance and \textbf{Bold} indicates the second best performance. The speed of hypergraph propagation (HP) in the reranking stage is 0.23s per 100 image pairs; it does not add any additional time or computing cost for pixel-level matching/segmentation. Compared with DELG+SP, OWL-ViT, SSP, and WarpCGLUNet, HD is much faster while maintaining high accuracy. Compared with D2R+Faster-RCNN+ASMK, HD achieves higher accuracy on both pixel retrieval and image retrieval (Table~\ref{tab:accuracy}). }
    \label{tab:pixel-map}
    \centering
    \small
    \setlength{\tabcolsep}{0pt}
    \newcolumntype{C}[1]{>{\centering\arraybackslash}p{#1}}
    \begin{tabular*}{\textwidth}{|C{0.102\textwidth}|C{0.28\textwidth}|C{0.0655\textwidth}|C{0.0655\textwidth}|C{0.0655\textwidth}|C{0.0655\textwidth}|C{0.0655\textwidth}|C{0.0655\textwidth}|C{0.0655\textwidth}|C{0.0655\textwidth}|C{0.0855\textwidth}|}
    \hline
    \multicolumn{2}{|c|}{\multirow{2}{*}{}} & \multicolumn{2}{c|}{PROxf} & \multicolumn{2}{c|}{PROxf+R1M} & \multicolumn{2}{c|}{PRPar} & \multicolumn{2}{c|}{PRPar+R1M} &\cellcolor{green}  \\ \cline{1-10}
         \multicolumn{2}{|l|}{}& M & H & M & H & M & H & M & H&\cellcolor{green}   \\ \cline{1-10}\cline{1-10}

        \multicolumn{10}{|c|}{  Image retrieval: DELG initial ranking~\cite{cao2020unifying} + HD reranking~\cite{an2021hypergraph}}&\multirow{-3}{*}{\cellcolor{green} \parbox{0.08\textwidth}{Overhead per 100 image pairs}} \\ \hline

        \multirow{6}{*}{ \makecell[c]{Pixel \\retrieval \\methods}}& DELG + SP~\cite{cao2020unifying} &\textbf{\textcolor{red}{39.6}}  &30.5  &\textbf{\textcolor{red}{36.0}} &28.2 &34.8  &20.2  &34.7 &19.5&41.22s  \\
         &D2R+Faster-RCNN+ASMK~\cite{teichmann2019detect} &30.1  &23.5 &30.5 &22.0 &26.3  &25.3  &25.7 &24.9& 0.11 s\\
         &OWL-VIT~\cite{minderer2022simple}  &12.3  &6.6 &12.1 &13.6 &7.9  & 7.6 &7.9 &7.8& 296.21s \\
         &SSP~\cite{fan2022ssp}  & \textbf{33.0} &29.7  &\textbf{35.7} &30.5 & \textbf{\textcolor{red}{46.4}} &\textbf{37.2}  &\textbf{\textcolor{red}{45.6}} &\textbf{37.2} & 62.33 s\\
         &WarpCGLUNet~\cite{truong2021warp}  & 31.2 &\textbf{32.6}  &31.5 &\textbf{31.7} &34.1  &27.3  & 34.3&28.1& 181.64s\\ 
        &HD &26.9  &\textbf{\textcolor{red}{39.7}}  &25.3&\textbf{\textcolor{red}{33.1}} &\textbf{43.7} &\textbf{\textcolor{red}{39.6}}  &\textbf{43.8} &\textbf{\textcolor{red}{38.9}}&0s \\ \hline
    
    \end{tabular*}

\end{table*}


\textbf{Baselines}.
Besides the image level performance, we compare our hypergraph diffusion with the state-of-the-art (SOTA) methods on pixel retrieval benchmarks PROxford and PRParis in Table~\ref{tab:pixel-map}. To find the pixel correspondences in the given database image, we choose the SOTA methods for landmark search, detection, segmentation, and dense matching fields. 
SPatial verification (SP) with Scale-Invariant Feature Transform (SIFT), DEep Local Feature (DELF), and DEep Local and Global features (DELG)~\cite{cao2020unifying} are the SOTA matching approaches in the landmark retrieval field. 
We also test the SOTA detection method Detect-to-Retrieval (D2R) in the landmark retrieval field, where the detailed implementation is explained in the next paragraph. 
The Vision Transformer for Open-World Localization
(OWL-ViT)~\cite{minderer2022simple} and One-Stage one-shot Detec-
tor (OS2D) are the SOTA methods in detection.
The Self-Support Prototype model (SSP)~\cite{fan2022ssp}, Hypercorrelation Squeeze Network (HSNet), and Mining model (Mining) are the SOTA segmentation methods.
The  Global-Local Universal Network (GLUNet), Probabilistic Dense Correspondence Network (PDCNet), and GLUNet trained using Globally Optimized Correspondence (GOCor) and Warp Consistency objective (WarpCGLUNet)~\cite{truong2021warp} are the SOTA methods in dense matching. We directly use the reported performance of these methods from the pixel retrieval benchmark~\cite{an2023towards}. 

We additionally introduce the implementation detail about D2R~\cite{teichmann2019detect} in Table~\ref{tab:miou} and~\ref{tab:pixel-map} because this is the first time to use its fine-tuned detector for localization purposes.
D2R was originally designed to increase the image-level retrieval performance; it detects many candidate regions for each database image to get a better global feature.  In this paper, we do not change the image-level ranking order for comparison fairness; Table~\ref{tab:accuracy} shows that our HD gives better results on the image-level ranking. For each proposed candidate region detected by D2R, we use the Aggregated Selective Match Kernel (ASMK) to aggregate all the DELG local features inside it. The region with the highest matching score with the ASMK feature of the query image is chosen as the final region. When aggregating the local features, we also tried the Vector of Locally Aggregated Descriptors (VLAD) and the simple average (Mean) aggregation.
We report its result of using Faster-RCNN~\cite{ren2015faster} or SSD~\cite{liu2016ssd} backbones fine-tuned on a huge manually boxed landmark dataset~\cite{teichmann2019detect}.

\textbf{Result}.
In Tables~\ref{tab:miou} and~\ref{tab:pixel-map}, we present a comparative analysis of pixel-retrieval accuracy between our Hypergraph Diffusion (HD) method and other state-of-the-art (SOTA) techniques. Unlike conventional approaches that compute direct correspondences for each query-database image pair, HD employs an innovative strategy. It diffuses spatial information from the query to the database images, which significantly cuts down on time and computational resources. Notably, despite initial expectations of a trade-off between speed and accuracy, our results reveal that HD's accuracy is not uniformly inferior to direct Spatial verification (SP) methods. Using the same DELG local features, HD demonstrates a reduced accuracy on the PROxford dataset but excels in the PRParis dataset compared to SP. 

The underlying reason for this performance differential might lie in our diffusion approach. Rather than matching query and database images directly via SP, HD initially identifies correspondences in more straightforward database images and subsequently tackles more challenging ones. This stepwise diffusion process effectively mitigates the impact of viewpoint and illumination variations between image pairs, breaking down severe changes into manageable phases.

To validate this hypothesis, we conducted visualizations, as depicted in Fig.~\ref{fig:pixel_retrieval_vis}. The visual evidence suggests that while the direct SP struggles to localize correspondence objects accurately in complex cases, our HD method achieves superior localization. It starts by detecting correspondences in less complex images, gradually extending to more challenging ones. Intriguingly, the `easier' database images contain additional visual cues absent in the query images and significantly aid the RANSAC-based SP in matching objects in more complex scenarios. For example, Fig.\ref{fig:pixel_retrieval_vis}-A illustrates how an  `easier' database image provides a clearer, unoccluded view of the target query landmark. Similarly, Fig.\ref{fig:pixel_retrieval_vis}-B shows that these database images offer viewpoints more closely aligned with the query image, aiding in object matching. In scenarios with changing illumination, as shown in Fig.~\ref{fig:pixel_retrieval_vis}-C, `easier' database images help mitigate this change, facilitating more accurate matching.
Furthermore, in Fig.~\ref{fig:pixel_retrieval_vis}-D, we observe that the target database image has a more complex background than the query image. Note that the input query is only the yellow region. This complexity often leads to outlier matchings in direct SP. However, the `easier' database images selected by our HD method provide additional background context, resulting in more accurate pixel retrieval. 


Tables~\ref{tab:miou} and~\ref{tab:pixel-map} illustrate that HD surpasses current SOTA detection methods capabilities and offers competitive, if not superior, performance than segmentation and dense-matching methods. 
Significantly, our approach merges image-level and pixel-level analyses, outpacing traditional two-step methods by delivering pixel-level results simultaneously with image-level rankings. This efficiency, crucially discussed in Section~\ref{subsec:time-memory}, positions HD as a robust and efficient solution for pixel retrieval tasks. It shows that shifting the spatial verification to an offline process not only enhances image-level performance but also drastically reduces retrieval time, achieving good pixel-retrieval performance. Considering accuracy and speed, HD emerges as a compelling choice for pixel retrieval.



\subsection{Effectivness of community selection}
\label{subsec:community selection result}

\begin{figure}[b]
\vspace{-20pt}
    \centering
    \includegraphics[width=0.45\textwidth, trim=50 270 50 380, clip]{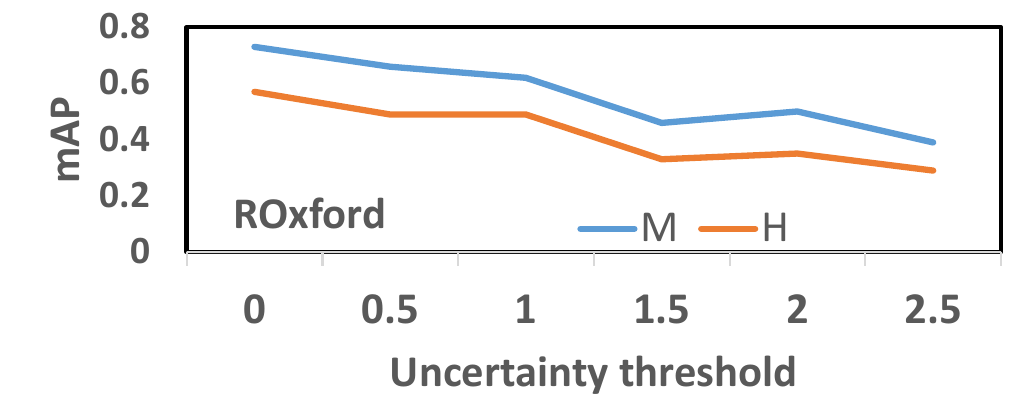}
    \includegraphics[width=0.45\textwidth, trim=50 270 50 380, clip ]{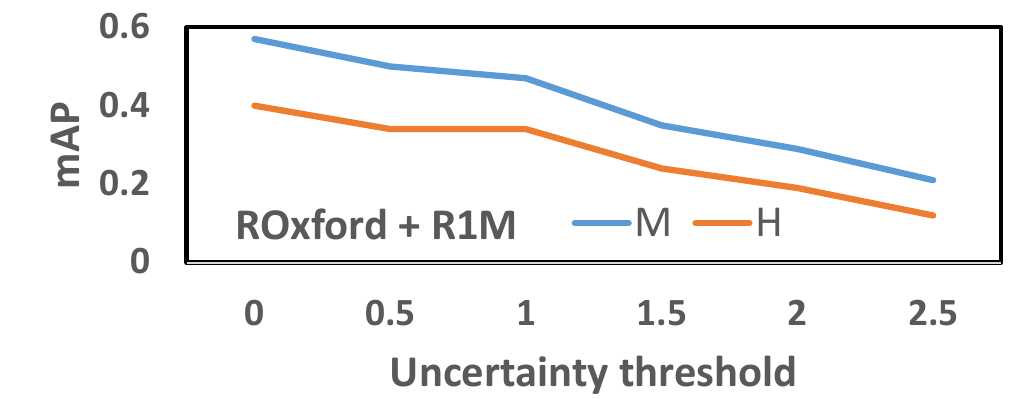} 
    \caption{mAP of the query results above the uncertainty threshold. Uncertainty predicts the query quality well.} 
    \label{fig:cs+mAP} 
  \end{figure} 

\begin{figure*}[h]
    \centering
    \includegraphics[width=\textwidth]{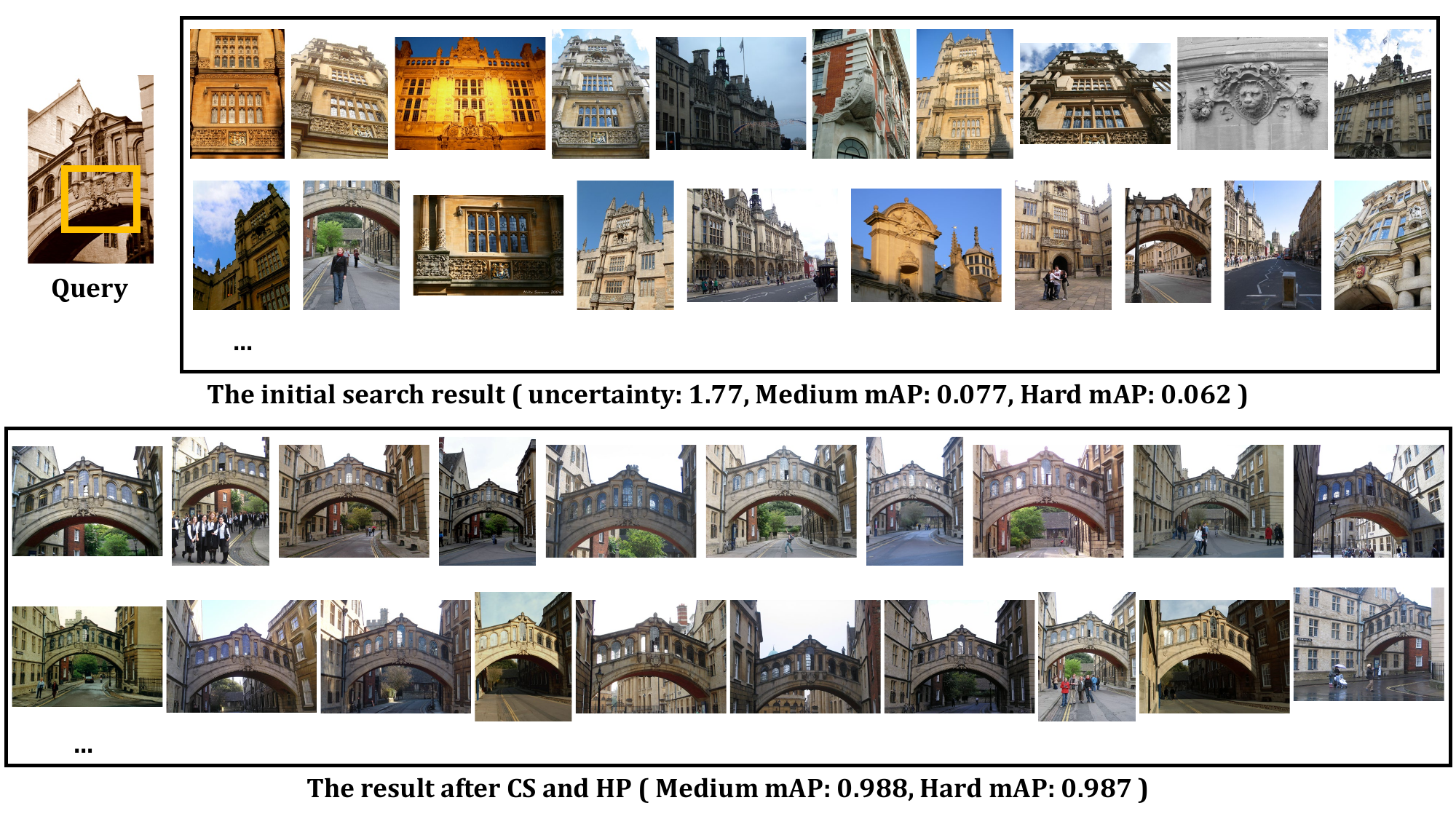}
    \caption{In this example, the top 11 images and the initial dominant community is unrelated to the query. Its uncertainty index is 1.77, which means the search engine is very uncertain about the accuracy of initial dominant community. So CS finds a new dominant community using spatial verification. After hypergraph propagation in this new dominant community, the retrieval performance is significantly improved.  
    }
    \label{fig:example}
\end{figure*}

\begin{table*}[b]
\tabcaption{
   Comparison between with and without community selection (CS) as the pre-stage when using spatial verification (SP) to initialize the hypergraph propagation (HP). This table shows both the mAP and the total number of queries which need SP when testing the corresponding dataset. 
    }
    \label{tab:CS} 
    \centering
    \small
    \newcolumntype{C}[1]{>{\centering\arraybackslash}p{#1}}
    \setlength{\tabcolsep}{0pt} 
    \begin{tabular*}{0.965\textwidth}{|C{0.16\textwidth}|C{0.16\textwidth}|C{0.16\textwidth}|C{0.16\textwidth}|C{0.16\textwidth}|C{0.16\textwidth}|}
    \hline
   
         &  HP &\multicolumn{2}{c|}{CS + SP + HP}  & \multicolumn{2}{c|}{SP + HP} \\ \hline
         & mAP & mAP & total \# SP & mAP & total \# SP \\ \hline
         ROxf(M/H) &85.7/70.3 &88.4/73.0 & \textbf{4} & 88.4/73.0 & 70  \\ \hline
         ROxf+R1M(M/H) &78.0/60.0 &79.1/60.5 &\textbf{7} &79.1 / 60.5& 70 \\  \hline
         RPar(M/H) &92.6/83.3 & 92.6/83.3 & \textbf{0} &  92.6/83.3 & 70\\ \hline
         RPar+R1M(M/H) &86.6/72.7 &86.6/72.7 &\textbf{0} &86.6/72.7 &70 \\ \hline 
    \end{tabular*}
    
\end{table*}

Figure~\ref{fig:cs+mAP} shows the mAP of the initial search results in ROxford with different uncertainty thresholds. As the uncertainty increases, the mAP of the initial search result decreases. The uncertainty index predicts the quality of the initial search without any user feedback. Only for the high uncertainty retrieval results, we use the computationally heavy approach, i.e., spatial verification, to improve the precision of nearest neighbors before the hypergraph propagation.  

Table~\ref{tab:CS} compares the performances between with and without community selection (CS) as the pre-stage when using spatial verification (SP) to initialize the hypergraph propagation (HP). With CS, the search engine improves the performance of hypergraph propagation in ROxford very efficiently by only conducting SP on 4 and 7 queries with and without R1M distractors, respectively. The performance improvement with R1M distractors is less than that without them. We think its reason is that the correct items with the R1M distractors are ranked outside the top 100 images and thus are not detected by SP. Although CS does not improve the retrieval results of these cases, it successfully tags them as uncertain cases.
We think this result leads to a promising future direction. Because once a search engine can automatically detect the failed query cases, it is possible to refine the representation ability by relabeling and training only for the failed cases instead of training on a large dataset. 

Thanks to the improved representation ability of DELG~\cite{cao2020unifying} features, the nearest neighbors of all queries in RParis dataset have high precision. The average precisions of the top 20 items of the initial result of all the queries in RParis are 0.97 and 0.90 in medium and hard protocols, respectively. As the initial dominant communities for all the queries in RParis are correct, online SP is not helpful for the propagation process. CS avoids these unnecessary SPs in these cases as the uncertainties of all the queries are low, as shown in Table~\ref{tab:CS}. 
CS significantly improves the retrieval performance for some hard cases. Figure~\ref{fig:example} shows an impressive example in ROxford, in which the mAP increases from 0.077 and 0.062 to 0.988 and 0.987 in medium and hard protocols, respectively. 

In summary, the main benefits of CS include: 1) CS quickly detects the low-quality retrieval result, which is either because of the difficult query or the insufficient feature representation ability; 2) CS avoids the unnecessary computationally heavy spatial verification process; 3) CS provides significant performance improvement for some hard cases.

\section{Time and memory cost}
\label{subsec:time-memory}

\begin{table*}[]
    \centering
     \newcolumntype{C}[1]{>{\centering\arraybackslash}p{#1}}
    \setlength{\tabcolsep}{0pt}
    \caption{Average data size per database image (in Byte), measured in Numpy array format. Spatial verification (SP) averagely requires 1040000 bytes, whereas Hypergraph Diffusion (HD) needs only 2678. HD demonstrates a significantly higher memory efficiency, being 388 times more efficient than SP.}
    \begin{tabular*}{\textwidth}{|C{0.33\textwidth}|C{0.337\textwidth}|C{0.33\textwidth}|}
    \hline
     Global feature & Matching information for HD &Local features for SP \\ \hline
    8192 B & 2678 B & 1040000 B  \\ \hline
    \end{tabular*}
\label{tab:memory}
\end{table*}

\begin{table*}[]
    \centering
     \newcolumntype{C}[1]{>{\centering\arraybackslash}p{#1}}
    \setlength{\tabcolsep}{0pt}
    \caption{Breakdown of average time per query.}
   \begin{tabular*}{\textwidth}{|C{0.33\textwidth}|C{0.337\textwidth}|C{0.33\textwidth}|}
    \hline
     Initial search & Hypergraph diffusion & Uncertainty calculation\\ \hline
      0.62 s & 1.07 s &0.0003 s  \\ \hline
    \end{tabular*}
\label{tab:time}
\end{table*}

\begin{table*}[h]
    \centering
     \newcolumntype{C}[1]{>{\centering\arraybackslash}p{#1}}
    \setlength{\tabcolsep}{0pt}
    \caption{Comparison of time complexity and latency across different pixel-retrieval approaches. K indicates the number
of images to calculate the pixel-level result. h (set as 3 in this paper) indicates the maximum iteration in hypergraph diffusion. r (set as 1000, the most common choice in this field) is the maximum RANSAC iteration time. l and d are the local features' numbers and dimensions, 1000 and 1024 for DELG. n is the image resolution or feature map size, f is the representation dimension, k is the kernel size of convolutions, and L is the number of Transformer or CNN layers. When testing the latency, we use an i9-9900K CPU @ 3.60GHz for HD and SP and a GeForce RTX™ 3090 Ti for deep learning methods. We use OWL~\cite{minderer2022simple} and WarpCGLUNet~\cite{truong2021warp} as the representative transformer-based and CNN-based models, respectively. Our method is much faster than the existing approaches for the pixel retrieval task.}
    \begin{tabular*}{\textwidth}{|C{0.196\textwidth}|C{0.2\textwidth}|C{0.2\textwidth}|C{0.2\textwidth}|C{0.2\textwidth}|}
    \hline
     Method &Hypergraph diffusion (ours) & Spatial verification & Transformer backbone & CNN backbone \\ \hline
      Time complexity per query& $O (h)$ & $O( K\times r \times l^2 \times d^2 )$    & $O(K\times L\times n^2 \times f )$ &$O(K\times L\times c\times n \times f^2 )$\\ \hline
       Latency per 100 image pairs&0.23s &41.22s  &296.2s (OWL~\cite{minderer2022simple}) &181.6s (WarpCGLUNet~\cite{truong2021warp})\\ \hline
    \end{tabular*}
\label{tab:complexity}
\end{table*}

In this section, we focus on assessing the time and memory costs associated with Hypergraph Diffusion (HD) in retrieval systems. While the accuracy of HD has been addressed in previous sections, here we aim to shed light on its efficiency and resource utilization, essential factors for practical deployment.

For hypergraph construction during query execution, spatial matching information for each database image and its k-nearest neighbors (with
k set to 200 as introduced in Sec.~\ref{subsec:implementation}) is recorded. Table~\ref{tab:memory} presents a comprehensive breakdown of memory usage. The matching information occupies an average of 2,678 bytes per database image. Notably, this memory usage constitutes only about 33\% of the memory required for the DELG global feature and a mere 0.2\% of that for local features. Note that HD does not need to load the local features in query time. This comparison illustrates HD's memory efficiency. 

In Table~\ref{tab:time}, we show the average query time during our experiments on the ROxford + R1M dataset. HD averages 1.07 s per query, while the initial search takes 0.62 s. This overhead, we argue, is reasonable given the benefits in accuracy and memory efficiency. Furthermore, the latency for uncertainty checks via our community selection technique is impressively minimal at 0.0003 s, bolstering our claim about the method's efficiency.

In Table~\ref{tab:complexity}, we compare the time complexity and latency of various pixel retrieval methods, including HD. Unlike Spatial Verification (SP) and neural network-based approaches, HD leverages pre-calculated matching information, leading to a time complexity linearly dependent on the diffusion process's maximum iteration number ($O(h)$). This simplicity starkly contrasts with the complexity of SP and the inference latency inherent in neural network methods, as shown in Table~\ref{tab:complexity}. To provide a tangible comparison, we document the latency experienced in processing 100 image pairs using different methods, with HD showing a marked advantage in speed. This comparative analysis underpins the practical viability of HD in time-sensitive applications. Our experiments used an i9-9900K CPU @ 3.60GHz for HD and SP and a GeForce RTX™ 3090 Ti for deep learning methods. This hardware setup reflects a high but realistic standard in computational resources, ensuring that our results represent potential real-world deployments.

\section{Discussion and conclusion}
\label{sec:discuss}

Our research introduces a novel approach to both image-level and pixel-level retrieval, leveraging hypergraph diffusion and community selection. This method has demonstrated high accuracy and remarkable efficiency in both time and memory usage, positioning it as a strong candidate for advanced retrieval systems. Despite these promising results, we acknowledge several areas for future exploration and improvement:

\textbf{Reducing Offline Computing Overhead}: Currently, our system relies on multiple CPUs for offline spatial verification, requiring approximately 12 hours to process the ROXford dataset on a single CPU. A transition to a GPU-based implementation is anticipated to enhance offline computing speed significantly.

\textbf{Enhancing Online Processing Speed}: The online processing speed, directly related to the number of diffusion steps in hypergraph diffusion, presents an opportunity for optimization. While our current implementation uses Python, transitioning to a C implementation could substantially improve processing speed, which is particularly important for real-time applications.

\textbf{Optimizing for Large-Scale Queries}: Improving the system's performance in large-scale query scenarios remains a critical challenge. Efficient handling of extensive datasets and queries is essential for practical applications.

\textbf{Database Management Efficiency}: An important aspect to consider is the efficient addition and deletion of images in the database. This process involves applying spatial verification for newly added database images, which requires efficient management to maintain system performance.

In conclusion, we believe that our approach combining hypergraph propagation with community selection takes a meaningful step in the realm of image and pixel search techniques. Notably, our method has achieved impressive mean Average Precision (mAP) scores of 73.0 and 60.5 on the ROxford dataset's hard protocol, with and without R1M distractors, respectively. To the best of our knowledge, these results represent state-of-the-art achievements in this field. We hope that our work lays a solid foundation for future research and development in efficient and accurate image retrieval systems.  

\section*{ Acknowledgement}
We express our profound appreciation to Prof. Giorgos Tolias from the Czech Technical University in Prague. His readiness to meticulously review drafts of our paper and offer comprehensive and constructive feedback has been instrumental to the advancement of our work.

{\small
\bibliographystyle{plain}
\bibliography{bare_jrnl_new_sample4}
}

\begin{IEEEbiography}[{\includegraphics[width=1in,height=1.25in,clip,keepaspectratio]{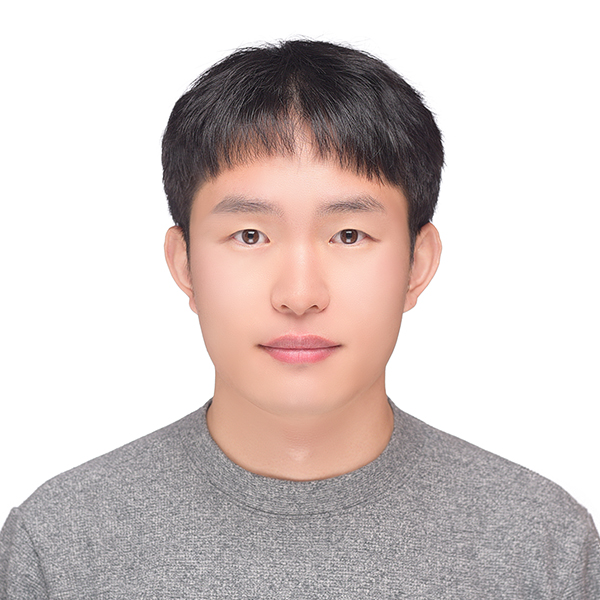}}]{Guoyuan An} is now a Ph.D. candidate in Computer Science at KAIST. He earned his B.S. degrees in Industrial Engineering and Computer Science from Inha University in 2018, followed by a M.S. degree in Computer Science from KAIST in 2020. His research focuses on large-scale image search, deep learning, and human-computer interaction.
\end{IEEEbiography}

\begin{IEEEbiography}[{\includegraphics[width=1in,height=1.25in,clip,keepaspectratio]{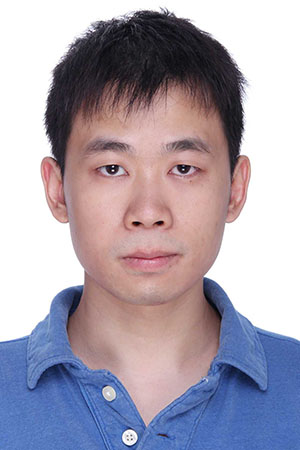}}]{Yuchi Huo} is an assistant professor under the ”Hundred Talent Program” in State Key Lab of CAD\&CG, Zhejiang University. His research interests are in rendering, deep learning, image search, and computational optics, which aim for the
realization of next-generation neural rendering pipelines and physical-neural computation.
\end{IEEEbiography}

\begin{IEEEbiography}[{\includegraphics[width=1in,height=1.25in,clip,keepaspectratio]{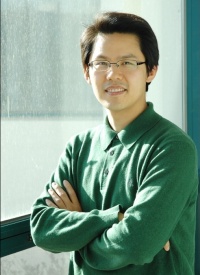}}]{Sung-Eui Yoon} (Senior Member, IEEE) received the
B.S. and M.S. degrees in computer science from Seoul National University, Seoul, South Korea, in
1999 and 2001, respectively, and the Ph.D. degree
in computer science from the University of North
Carolina at Chapel Hill, Chapel Hill, NC, USA, in
2005. He is currently a Professor with the Korea Advanced Institute of Science and Technology, Daejeon, South Korea. He was a Postdoctoral Scholar
with Lawrence Livermore National Laboratory, Livermore, CA, USA. He has authored or coauthored
more than 70 technical papers in top journals and conference related to graphics, vision, and robotics, a monograph on real-time massive model rendering
with other three coauthors in 2008, and recently an online book on Rendering in
2018. His research interests include rendering, image search, and motion planning spanning graphics, vision, and robotics. He also gave numerous tutorials
on ray tracing, collision detection, image search, and sound source localization
in premier conferences like ACM SIGGRAPH, IEEE Visualization, CVPR, and
ICRA. He was the conference Co-chair and Paper Co-chair for ACM I3D 2012
and 2013 respectively. Some of his papers was the recipient of the Test-of-Time
Award, a distinguished Paper Award, and a few invitations to IEEE Transaction
on Visualization and Graphics. He is a Senior Member of ACM.
\end{IEEEbiography}

\vfill

\end{document}